\documentclass[11pt,a4paper]{article}
\usepackage[hyperref]{naaclhlt2018}
\usepackage{times}
\usepackage{latexsym}

\usepackage[T2A, T1]{fontenc}
\usepackage[utf8]{inputenc}

\usepackage{url}
\usepackage{graphicx}
\usepackage{multirow}
\usepackage{amsthm}
\usepackage{amssymb}
\usepackage{amsmath}
\usepackage{amsfonts}
\usepackage{color}
\usepackage{mdwlist}
\usepackage{microtype}
\usepackage{tabularx}
\usepackage{booktabs}
\usepackage{subcaption}
\usepackage{mwe} %<- For dummy image
\DeclareCaptionLabelFormat{andtable}{#1˜#2 \& \tablename˜\thetable}

\usepackage{color,colortbl}

\definecolor{Gray}{gray}{0.9}
\usepackage{enumitem,amssymb}
\newlist{todolist}{itemize}{2}
\setlist[todolist]{label=$\square$}
\usepackage{pifont}
\newcolumntype{b}{X}
\newcolumntype{m}{>{\hsize=.6\hsize}X}
\newcolumntype{s}{>{\hsize=.33\hsize}X}

\makeatletter
\newcommand{\removelatexerror}{\let\@latex@error\@gobble}

\makeatother

\usepackage[russian,english]{babel}

\aclfinalcopy % Uncomment this line for the final submission
\title{Post-Specialisation: Retrofitting Vectors of Words \\ Unseen in Lexical Resources}

\author{Ivan Vuli\'c$^{\mathbf{1}}$, ~ Goran Glava\v{s}$^{\mathbf{2}}$, ~ Nikola Mrk\v{s}i\'{c}$^{\mathbf{3}}$, ~ {Anna Korhonen}$^{\mathbf{1}}$\\
$^{\mathbf{1}}$ Language Technology Lab, University of Cambridge \\
$^{\mathbf{2}}$ Data and Web Science Group, University of Mannheim \\
$^{\mathbf{3}}$ PolyAI\\
\texttt{\{iv250,alk23\}@cam.ac.uk} \\ \texttt{goran@informatik.uni-mannheim.de} \hspace{0.8em} \texttt{nikola@poly-ai.com} }

\date{}

\begin{document}
\maketitle
\begin{abstract}
Word vector \textit{specialisation} (also known as \emph{retrofitting}) is a portable, light-weight approach to fine-tuning {arbitrary} distributional word vector spaces by \emph{injecting} external knowledge from rich lexical resources such as WordNet. By design, these \emph{post-processing} methods only update the vectors of words occurring in external lexicons, leaving the representations of all \textit{unseen} words intact. In this paper, we show that constraint-driven vector space specialisation can be extended to unseen words. We propose a novel  \textit{post-specialisation} method that: \textbf{a)} preserves the useful linguistic knowledge for seen words; {while} \textbf{b)} propagating this external signal to unseen words in order to improve their vector representations as well. Our {post-specialisation} approach explicits a non-linear specialisation function in the form of a deep neural network by learning to predict specialised vectors from their original distributional counterparts. The learned function is then used to specialise vectors of unseen words.    
%Our {post-specialisation} method learns a non-linear function based on a deep neural network which maps unseen words from the initial distributional space to the specialised space. 
This approach, applicable to any post-processing model, yields considerable gains over the initial specialisation models both in intrinsic word similarity tasks, and in two downstream tasks: \emph{dialogue state tracking} and \textit{lexical text simplification}. The positive effects persist across three languages, demonstrating the importance of specialising the full vocabulary of distributional word vector spaces.
\end{abstract}

\section{Introduction}
\label{s:intro}
Word representation learning is a key research area in current Natural Language Processing (NLP), with its usefulness demonstrated across a range of tasks \cite{Collobert:2011jmlr,Chen:2014emnlp,Melamud:2016naacl}. The standard techniques for inducing distributed word representations are grounded in the distributional hypothesis \cite{Harris:1954}: they rely on co-occurrence information in large textual corpora \cite{Mikolov:2013nips,Pennington:2014emnlp,Levy:2014acl,Levy:2015tacl,Bojanowski:2017tacl}. As a result, these models tend to coalesce the notions of semantic similarity and (broader) conceptual relatedness, and cannot accurately distinguish antonyms from synonyms \cite{Hill:2015cl,Schwartz:2015conll}. Recently, we have witnessed a rise of interest in representation models that move beyond stand-alone unsupervised learning: they leverage external knowledge in human- and automatically-constructed lexical resources to enrich the semantic content of distributional word vectors, in a process termed \textit{semantic specialisation}.

This is often done as a post-processing (sometimes referred to as \textit{retrofitting}) step: input word vectors are \textit{fine-tuned} to satisfy linguistic constraints extracted from lexical resources such as WordNet or BabelNet \cite{Faruqui:2015naacl,Mrksic:2017tacl}. The use of external curated knowledge yields improved word vectors for the benefit of downstream applications \cite{Faruqui:2016thesis}. At the same time, this specialisation of the distributional space distinguishes between true similarity and relatedness, and supports language understanding tasks \cite{Kiela:2015emnlp,Mrksic:2017acl}.

While there is consensus regarding their benefits and ease of use, one property of the post-processing specialisation methods slips under the radar: most existing post-processors update word embeddings only for words which are present (i.e., \textit{seen}) in the external constraints, while vectors of all other (i.e., \textit{unseen}) words remain unaffected. In this work, we propose a new approach that extends the specialisation framework to unseen words, relying on the transformation of the vector (sub)space of seen words. Our intuition is that the process of fine-tuning seen words provides implicit information on how to leverage the external knowledge to unseen words. The method should preserve the already injected knowledge for seen words, simultaneously propagating the external signal to unseen words in order to improve their vectors.

The proposed \textit{post-specialisation} method can be seen as a two-step process, illustrated in Fig.~\ref{fig:main1}: \textbf{1)} We use a state-of-the-art specialisation model to transform the subspace of seen words from the input distributional space into the specialised subspace; \textbf{2)} We learn a mapping function based on the transformation of the ``seen subspace'', and then apply it to the distributional subspace of unseen words. We allow the proposed post-specialisation model to learn from large external linguistic resources by implementing the mapping as a deep feed-forward neural network with non-linear activations. This allows the model to learn the generalisation of the fine-tuning steps taken by the initial specialisation model, itself based on a very large number (e.g., hundreds of thousands) of external linguistic constraints.

As indicated by the results on word similarity and two downstream tasks (dialogue state tracking and lexical text simplification) our post-specialisation method consistently outperforms state-of-the-art methods which specialise seen words only. We report improvements using three distinct input vector spaces for English and for three test languages (English, German, Italian), verifying the robustness of our approach.

\section{Related Work and Motivation}
\label{s:related}
\paragraph{Vector Space Specialisation} 
A standard approach to incorporating external and background knowledge into word vector spaces is to pull the representations of similar words closer together and to push words in undesirable relations (e.g., antonyms) away from each other. Some models integrate such constraints into the training procedure and \textit{jointly} optimize distributional and non-distributional objectives: they modify the prior or the regularisation \cite{Yu:2014,Xu:2014,Bian:14,Kiela:2015emnlp}, or use a variant of the SGNS-style objective \cite{Liu:EtAl:15,Ono:2015naacl,Osborne:16,Nguyen:2017emnlp}. In theory, word embeddings obtained by these joint models could be as good as representations produced by models which fine-tune input vector space. However, their performance falls  behind that of fine-tuning methods \cite{Wieting:2015tacl}. Another disadvantage is that their architecture is tied to a specific underlying model (typically \texttt{word2vec} models).

In contrast, fine-tuning models inject external knowledge from available lexical resources (e.g., WordNet, PPDB) into pre-trained word vectors as a \textit{post-processing step} \cite{Faruqui:2015naacl,Rothe:2015acl,Wieting:2015tacl,Nguyen:2016acl,Mrksic:2016naacl,Cotterell:2016acl,Mrksic:2017tacl}. Such post-processing models are popular because they offer a portable, flexible, and light-weight approach to incorporating external knowledge into \textit{arbitrary} vector spaces, yielding state-of-the-art results on language understanding tasks \cite{Faruqui:2015naacl,Mrksic:2016naacl,Kim:2016slt,Vulic:2017acl}. 

Existing post-processing models, however, suffer from a major limitation. Their \textit{modus operandi} is to enrich the distributional information with external knowledge only if such knowledge is present in a lexical resource. This means that they update and improve only representations of words actually {seen} in external resources. Because such words constitute only a fraction of the whole vocabulary (see Sect.~\ref{s:exp}), most words, {unseen} in the constraints, retain their original vectors. The main goal of this work is to address this shortcoming by specialising \textit{all} words from the initial distributional space. %

\section{Methodology: Post-Specialisation} 
\label{s:methodology}
Our starting point is the state-of-the-art specialisation model \textsc{attract-repel (ar)} \cite{Mrksic:2017tacl}, outlined in Sect.~\ref{ss:ar}. We opt for the \textsc{ar} model due to its strong performance and ease of use, but we note that the proposed post-specialisation approach for specialising unseen words, described in Sect.~\ref{ss:unseen}, is applicable to any post-processor, as empirically validated in Sect.~\ref{s:results}.

\subsection{Initial Specialisation Model: \textsc{ar}}
\label{ss:ar}
Let $\mathcal{V}_s$ be the vocabulary, $A$ the set of synonymous \textsc{attract} word pairs (e.g.,~\emph{rich} and \emph{wealthy}), and $R$ the set of antonymous \textsc{repel} word pairs (e.g.,~\emph{increase} and \emph{decrease}). The \textsc{attract-repel} procedure operates over mini-batches of such pairs $\mathcal{B}_{A}$ and $\mathcal{B}_{R}$. Let each word pair $(x_l, x_r)$ in these sets correspond to a vector pair $(\mathbf{x}_l, \mathbf{x}_r)$. A mini-batch of $b_{att}$ \text{attract} word pairs is given by $\mathcal{B}_{A} = \lbrack (\mathbf{x}_{l}^{1}, \mathbf{x}_{r}^{1}), \ldots, (\mathbf{x}_{l}^{k_{1}}, \mathbf{x}_{r}^{k_1})\rbrack$ (analogously for $\mathcal{B}_{R}$, which consists of $b_{rep}$ pairs).

Next, the sets of negative examples $T_{A} = \lbrack (\mathbf{t}_{l}^{1}, \mathbf{t}_{r}^{1}), \ldots, (\mathbf{t}_{l}^{k_{1}}, \mathbf{t}_{r}^{k_1})\rbrack$ and $T_{R} = \lbrack (\mathbf{t}_{l}^1, \mathbf{t}_{r}^1), \ldots, (\mathbf{t}_{l}^{k_2}, \mathbf{t}_{r}^{k_2})\rbrack $ are defined as pairs of \emph{negative examples} for each $A$ and $R$ pair in mini-batches $\mathcal{B}_{A}$ and $\mathcal{B}_{R}$. These negative examples are chosen from the word vectors present in $\mathcal{B}_A$ or $\mathcal{B}_R$ so that, for each $A$ pair $(\mathbf{x}_l, \mathbf{x}_r)$, the negative example pair $(\mathbf{t}_l, \mathbf{t}_r)$ is chosen so that $\mathbf{t}_l$ is the vector closest (in terms of cosine distance) to $\mathbf{x}_l$ and $\mathbf{t}_r$ is closest to $\mathbf{x}_r$.\footnote{Similarly, for each $R$ pair $(\mathbf{x}_l, \mathbf{x}_r)$, the negative pair $(\mathbf{t}_l, \mathbf{t}_r)$ is chosen from the in-batch vectors so that $\mathbf{t}_l$ is the vector furthest away from $\mathbf{x}_l$ and $\mathbf{t}_r$ is furthest from $\mathbf{x}_r$. All vectors are unit length (re)normalised after each epoch.} The negatives are used \textbf{1)} to force $A$ pairs to be closer to each other than to their respective negative examples; and \textbf{2)} to force $R$ pairs to be further away from each other than from their negative examples. The first term of the cost function pulls $A$ pairs together:
%\vspace{-0.25cm}

\vspace{-0em}
{\footnotesize
\begin{align} 
Att(\mathcal{B}_A, T_{A}) ~=  \sum_{ i = 1}^{b_{att}} & \big[~ \tau \left( \delta_{att} +  \mathbf{x}_l^{i} \mathbf{t}_l^{i} - \mathbf{x}_l^{i} \mathbf{x}_r^{i} \right) \notag \\
+& \tau \left( \delta_{att} +  \mathbf{x}_r^{i} \mathbf{t}_r^{i} - \mathbf{x}_l^{i} \mathbf{x}_r^{i}  \right) \big]
\end{align}}%
\noindent where $\tau(z)=\max(0,z)$ is the standard rectifier function \cite{Nair:2010icml} and $\delta_{att}$ is the attract margin: it determines how much closer these vectors should be to each other than to their respective negative examples. The second, \textsc{repel} term in the cost function is analogous: it pushes $R$ word pairs away from each other by the margin $\delta_{rep}$. 

Finally, in addition to the $A$ and $R$ terms, a regularisation term is used to preserve the semantic content originally present in the distributional vector space, as long as this information does not contradict the injected external knowledge. Let $\mathcal{V}(\mathcal{B})$ be the set of all word vectors present in a mini-batch, the distributional regularisation term is then:

\vspace{-0em}
{\footnotesize
\begin{align}
  Reg(\mathcal{B}_A, \mathcal{B}_R) =  \sum\limits_{ \mathbf{x}_i \in V(\mathcal{B}_A \cup \mathcal{B}_R) }  \lambda_{reg} \left\| \widehat{\mathbf{x}_{i}} - \mathbf{x}_i \right\|_{2} 
 \end{align}}%
\noindent where $\lambda_{reg}$ is the $L_2$-regularisation constant and $\widehat{\mathbf{x}_{i}}$ denotes the original (distributional) word vector for word $x_i$. The full \textsc{attract-repel} cost function is finally constructed as the sum of all three terms.

%% PROBLEM ILLUSTRATION
\begin{figure*}[t]
    \centering
    \begin{subfigure}[t]{0.448\linewidth}
        \centering
        \includegraphics[width=1.00\linewidth]{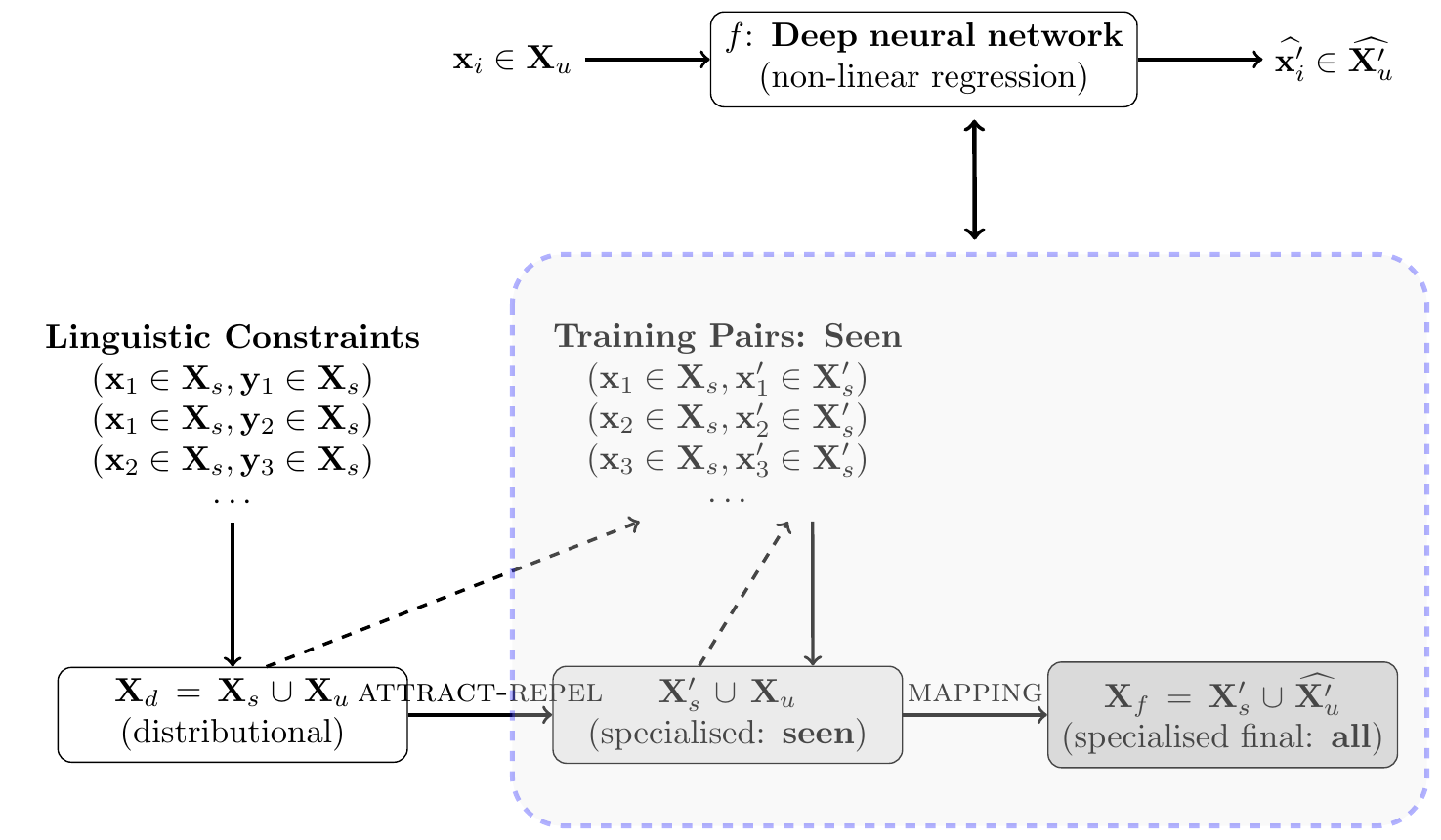}
        \vspace{0.2em}
        \caption{High-level \textbf{illustration}}
        \label{fig:main1}
    \end{subfigure}
    \begin{subfigure}[t]{0.535\textwidth}
        \centering
        \includegraphics[width=1.00\linewidth]{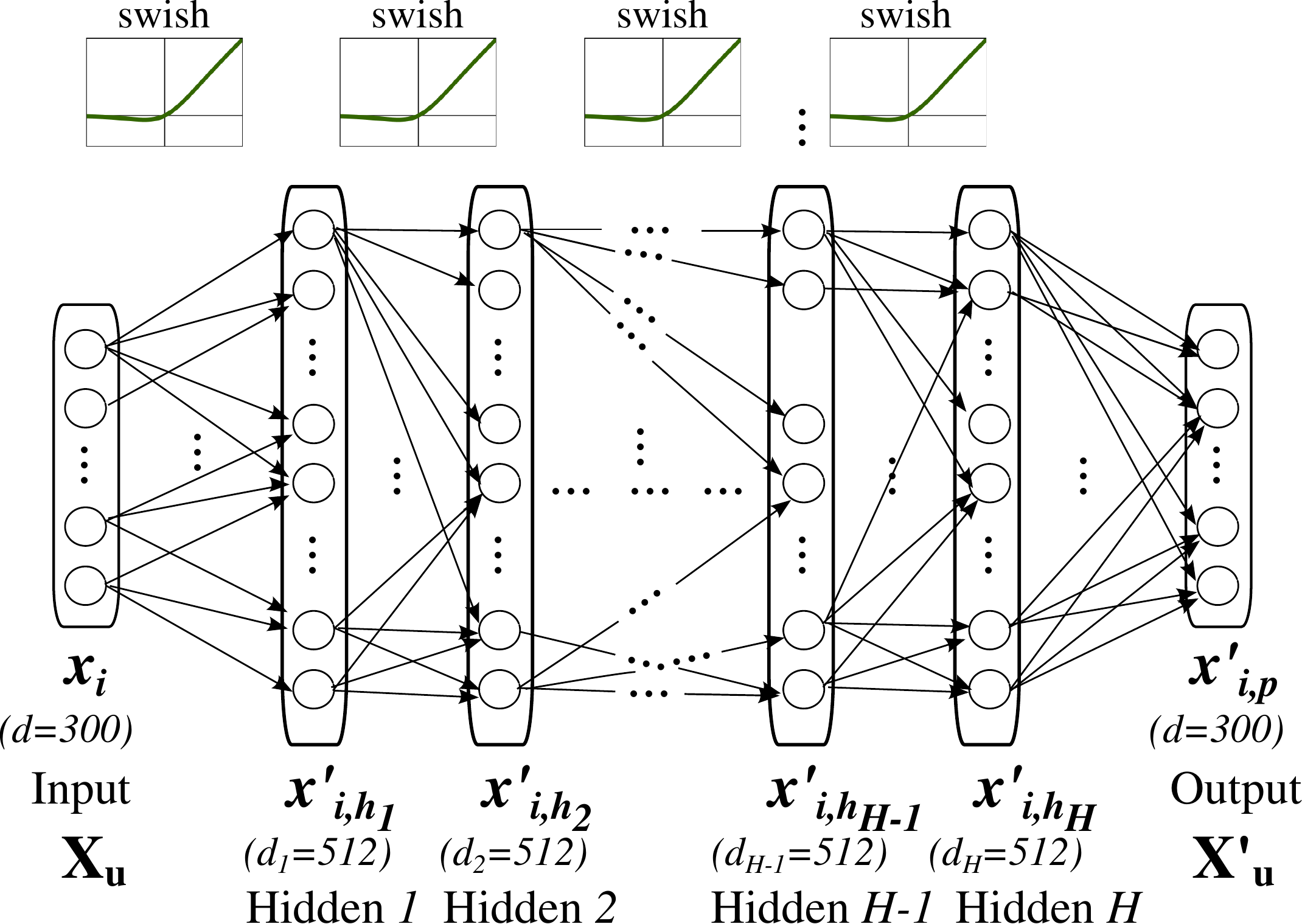}
        \vspace{0.2em}
        \caption{Low-level \textbf{implementation}: deep feed-forward neural network}
        \label{fig:main2}
    \end{subfigure}
    \vspace{-1mm}
    \caption{\textbf{(a)} High-level illustration of the post-specialisation approach: the subspace $\mathbf{X}_s$ of the initial distributional vector space $\mathbf{X}_d = \mathbf{X}_s \cup \mathbf{X}_u$ is first specialised/fine-tuned by the \textsc{attract-repel} specialisation model (or any other post-processing model) to obtain the transformed subspace $\mathbf{X}'_s$. The words present (i.e., \textit{seen}) in the input set of linguistic constraints are now assigned different representations in $\mathbf{X}_s$ (the original distributional vector) and $\mathbf{X}'_s$ (the specialised vector): they are therefore used as training examples to learn a non-linear cross-space mapping function. This function is then applied to all word vectors $\mathbf{x}_i \in \mathbf{X}_u$ representing words \textit{unseen} in the constraints to yield a specialised subspace $\widehat{\mathbf{X}'_u}$. The final space is $\mathbf{X}_f = \mathbf{X}'_s \cup \widehat{\mathbf{X}'_u}$, and it contains transformed representations for \textit{all} words from the initial space $\mathbf{X}_d$. \textbf{(b)} The actual implementation of the non-linear regression function which maps from $\mathbf{X}_u$ to $\widehat{\mathbf{X}'_u}$: a deep feed-forward fully-connected neural net with non-linearities and $H$ hidden layers.}
\vspace{-0.5mm}
\label{fig:main}
\end{figure*}

% (IV, RemoveD): The final vector space $\mathbf{X}_f$ now contains transformed representations for \textit{all} words from the initial space $\mathbf{X}_d$

\subsection{Specialisation of Unseen Words}
\label{ss:unseen}

\paragraph{Problem Formulation} 
The goal is to learn a \textit{global transformation function} that generalises the perturbations of the initial vector space made by \textsc{attract-repel} (or any other specialisation procedure), as conditioned on the external constraints. The learned function propagates the signal coded in the input constraints to all the words unseen during the specialisation process. We seek a regression function $f: \mathbb{R}^{dim} \rightarrow \mathbb{R}^{dim}$, where $dim$ is the vector space dimensionality. It maps word vectors from the initial vector space $\mathbf{X}$ to the specialised target space $\mathbf{X}'$. Let $\widehat{\mathbf{X}'} = f(\mathbf{X})$ refer to the predicted mapping of the vector space, while the mapping of a single word vector is denoted $\widehat{\mathbf{x}'_{i}} = f(\mathbf{x}_i)$.

%\paragraph{Terminology and Notation}
An input distributional vector space $\mathbf{X}_d$ represents words from a vocabulary $\mathcal{V}_d$. $\mathcal{V}_d$ may be divided into two vocabulary subsets: $\mathcal{V}_d = \mathcal{V}_s \cup \mathcal{V}_{u}$, $\mathcal{V}_s \cap \mathcal{V}_{u} = \emptyset$, with the accompanying vector subspaces $\mathbf{X}_d = \mathbf{X}_s \sqcup \mathbf{X}_{u}$. $\mathcal{V}_s$ refers to the vocabulary of {seen words}: those that appear in the external linguistic constraints and have their embeddings changed in the specialisation process. $\mathcal{V}_u$ denotes the vocabulary of {unseen words}: those not present in the constraints and whose embeddings are unaffected by the specialisation procedure.

The \textsc{ar} specialisation process transforms only the subspace $\mathbf{X}_s$ into the specialised subspace $\mathbf{X}'_s$. All words $x_i \in \mathcal{V}_s$ may now be used as training examples for learning the explicit mapping function $f$ from $\mathbf{X}_s$ into $\mathbf{X}'_s$. If $N=|\mathcal{V}_s|$, we in fact rely on $N$ training pairs: $(\mathbf{x}_i, \mathbf{x}'_i) = \{\mathbf{x}_i \in \mathbf{X}_s, \mathbf{x}'_i \in \mathbf{X}'_s\}$. Function $f$ can then be applied to unseen words $x \in \mathcal{V}_u$ to yield the specialised subspace $\widehat{\mathbf{X}'_u} = f(\mathbf{X}_u)$. The specialised space containing \textit{all} words is then $\mathbf{X}_f = \mathbf{X}'_s \cup \widehat{\mathbf{X}'_u}$. The complete high-level post-specialisation procedure is outlined in Fig.~\ref{fig:main1}.

Note that another variant of the approach could obtain $\mathbf{X}_f$ as $\mathbf{X}_f = f(\mathbf{X}_d)$, that is, the entire distributional space is transformed by $f$. However, this variant seems counter-intuitive as it forgets the actual output of the initial specialisation procedure and replaces word vectors from $\mathbf{X}'_s$ with their approximations, i.e., $f$-mapped vectors.\footnote{We have empirically confirmed the intuition that the first variant is superior to this alternative. We do not report the actual quantitative comparison for brevity.}

%\vspace{1.4mm}
\paragraph{Objective Functions} 
As mentioned, the $N$ \textit{seen} words $x_i \in \mathcal{V}_s$ in fact serve as our ``pseudo-translation'' pairs supporting the learning of a cross-space mapping function. In practice, in its high-level formulation, our mapping problem is equivalent to those encountered in the literature on cross-lingual word embeddings where the goal is to learn a shared cross-lingual space given monolingual vector spaces in two languages and $N_1$ translation pairs \cite{Mikolov:2013arxiv,Lazaridou:2015acl,Vulic:2016acl,Artetxe:2016emnlp,Artetxe:2017acl,Conneau:2017arxiv,Ruder:2017arxiv}. In our setup, the standard objective based on $L_2$-penalised least squares may be formulated as follows:

\vspace{-0.0em}
 {\footnotesize
 \begin{align}
f_{\textsc{mse}} = \text{arg min}||f(\mathbf{X}_s) - \mathbf{X}'_s||^2_F
 \end{align}}%
where $||\cdot||_F^2$ denotes the squared Frobenius norm. In the most common form $f(\mathbf{X}_s)$ is simply a linear  map/matrix $\mathbf{W}_f \in \mathbb{R}^{dim \times dim}$ \cite{Mikolov:2013arxiv} as follows: $f(\mathbf{X}) = \mathbf{W}_f\mathbf{X}$.

%\vspace{-0.8em}
% {\footnotesize
% \begin{align}
%f(\mathbf{X}) = \mathbf{W}_f\mathbf{X}
 %\end{align}}%
After learning $f$ based on the $\mathbf{X}_s \rightarrow \mathbf{X}'_s$ transformation, one can simply apply $f$ to unseen words: $\widehat{\mathbf{X}'_u} = f(\mathbf{X}_u)$. This linear mapping model, termed \textsc{linear-mse}, has an analytical solution \cite{Artetxe:2016emnlp}, and has been proven to work well with cross-lingual embeddings. However, given that the specialisation model injects hundreds of thousands (or even millions) of linguistic constraints into the distributional space (see later in Sect.~\ref{s:exp}), we suspect that the assumption of linearity is too limiting and does not fully hold in this particular setup.

Using the same $L_2$-penalized least squares objective, we can thus replace the linear map with a \textit{non-linear} function $f: \mathbb{R}^{dim} \rightarrow \mathbb{R}^{dim}$. The non-linear mapping, illustrated by Fig.~\ref{fig:main2}, is implemented as a deep feed-forward fully-connected neural network (DFFN) with $H$ hidden layers and non-linear activations. This variant is called \textsc{nonlinear-mse}.

Another variant objective is the contrastive margin-based ranking loss with negative sampling (\textsc{mm}) similar to the original \textsc{attract-repel} objective, used in other applications in prior work (e.g., for cross-modal mapping) \cite{Weston:2011ijcai,Frome:2013nips,Lazaridou:2015acl,Kummerfeld:2015emnlp}. Let $\widehat{\mathbf{x'}_{i}} = f(\mathbf{x_i})$ denote the predicted vector for the word $x_i \in \mathcal{V}_s$, and let $\mathbf{x'}_{i}$ refer to the ``true'' vector of $x_i$ in the specialised space $\mathbf{X}'_s$ after the \textsc{ar} specialisation procedure. The \textsc{mm} loss is then defined as follows:

 \vspace{-0.0em}
 {\footnotesize
 \begin{align}
 J_{\textsc{MM}} = \sum_{i=1}^N \sum_{j \neq i}^k \tau\Big(\delta_{mm} - cos\big(\widehat{\mathbf{x'}_{i}},\mathbf{x'}_{i}\big) + cos \big(\widehat{\mathbf{x'}_{i}},\mathbf{x'}_{j}\big)\Big) \notag
 \label{eq:maxmargin}
 \end{align}}%
where $cos$ is the cosine similarity measure, $\delta_{mm}$ is the margin, and $k$ is the number of negative samples. The objective tries to learn the mapping $f$ so that each predicted vector $\widehat{\mathbf{x'}_{i}}$ is by the specified margin $\delta_{mm}$ closer to the correct target vector $\mathbf{x'}_{i}$ than to any other of $k$ target vectors $\mathbf{x'}_{j}$ serving as negative examples.\footnote{We have also experimented with a simpler \textit{hinge loss} function without negative examples, formulated as $J = \sum_{i=1}^N \tau\Big(\delta_{mm} - cos(\widehat{\mathbf{x'}_{i}},\mathbf{x'}_{i}) \Big)$. For instance, with $\delta_{mm} = 1.0$ the idea is to learn a mapping $f$ that, for each $x_i$ enforces the predicted vector and the correct target vector to have a maximum cosine similarity. We do not report the results with this variant as, although it outscores the \textsc{mse}-style objective, it was consistently outperformed by the \textsc{mm} objective.} Function $f$ can again be either a simple linear map (\textsc{linear-mm}), or implemented as a DFFN (\textsc{nonlinear-mm}, see Fig.~\ref{fig:main2}).

\section{Experimental Setup}
\label{s:exp}
\noindent \textbf{Starting Word Embeddings ($\mathbf{X}_d = \mathbf{X}_s \cup \mathbf{X}_u$)} 
To test the robustness of our approach, we experiment with three well-known, publicly available collections of English word vectors: \textbf{1)} Skip-Gram with Negative Sampling (\textsc{sgns-bow2}) \cite{Mikolov:2013nips} trained on the Polyglot Wikipedia \cite{AlRfou:2013conll} by \newcite{Levy:2014acl} using bag-of-words windows of size 2; \textbf{2)} \textsc{glove} Common Crawl \cite{Pennington:2014emnlp}; and \textbf{3)} \textsc{fastText} \cite{Bojanowski:2017tacl}, a SGNS variant which builds word vectors as the sum of their constituent character n-gram vectors. All word embeddings are $300$-dimensional.\footnote{For further details regarding the architectures and training setup of the used vector collections, we refer the reader to the original papers. Additional experiments with other word vectors, e.g., with \textsc{context2vec} \cite{Melamud:2016conll} (which uses bidirectional LSTMs \cite{Hochreiter:1997} for context modeling), and with dependency-word based embeddings \cite{Bansal:2014acl,Melamud:2016naacl} lead to similar results and same conclusions.}

%\vspace{1.6mm}
\paragraph{AR Specialisation and Constraints ($\mathbf{X}_s \rightarrow \mathbf{X}'_s$)} 
We experiment with linguistic constraints used before by \cite{Mrksic:2017tacl,Vulic:2017emnlp}: they extracted monolingual synonymy/\textsc{attract} pairs from the Paraphrase Database (PPDB) \cite{Ganitkevitch:2013naacl,Pavlick:2015acl} (640,435 synonymy pairs in total), while their antonymy/\textsc{repel} constraints came from BabelNet \cite{Navigli:12} (11,939 pairs).\footnote{We have experimented with another set of constraints used in prior work \cite{Zhang:2014emnlp,Ono:2015naacl}, reaching similar conclusions: these were extracted from WordNet \cite{Fellbaum:1998wn} and Roget \cite{Kipfer:2009book}, and comprise 1,023,082 synonymy pairs and 380,873 antonymy pairs.}

% We observe similar trends in results with this set of constraints.

%\footnote{https://github.com/tticoin/AntonymDetection} 

%Unless stated otherwise, all experiments are conducted using \textsc{const-1}.

The coverage of $\mathcal{V}_d$ vocabulary words in the constraints illustrates well the problem of unseen words with the fine-tuning specialisation models. For instance, the constraints cover only a small subset of the entire vocabulary $\mathcal{V}_d$ for \textsc{sgns-bow2}: 16.6\%. They also cover only 15.3\% of the top 200K most frequent $\mathcal{V}_d$ words from \textsc{fastText}.

%\vspace{1.6mm}
\paragraph{Network Design and Parameters ($\mathbf{X}_u \rightarrow \widehat{\mathbf{X}'_u}$)} 
The non-linear regression function $f: \mathbb{R}^d \rightarrow \mathbb{R}^d$ is a DFFN with $H$ hidden layers, each of dimensionality $d_1=d_2=\ldots=d_{H}=512$ (see Fig.~\ref{fig:main2}). Non-linear activations are used in each layer and omitted only before the final output layer to enable full-range predictions (see Fig.~\ref{fig:main2} again).

The choices of non-linear activation and initialisation are guided by recent recommendations from the literature. First, we use \textit{swish} \cite{Swish:2017arxiv,Elfwing:2017arxiv} as non-linearity, defined as $swish(x)=x \cdot \text{sigmoid}(\beta x)$. We fix $\beta=1$ as suggested by \newcite{Swish:2017arxiv}.\footnote{According to \newcite{Swish:2017arxiv}, for deep networks \textit{swish} has a slight edge over the family of LU/ReLU-related activations \cite{Maas:2014icml,He:2015iccv,Klambauer:2017arxiv}. We also observe a minor (and insignificant) difference in performance in favour of \textit{swish}.} Second, we use the \textsc{he} normal initialisation \cite{He:2015iccv}, which is preferred over the \textsc{xavier} initialisation \cite{Glorot:2010aistats} for deep models \cite{Mishkin:2016iclr,Li:2016arxiv}, although in our experiments we do not observe a significant difference in performance between the two alternatives. We set $H=5$ in all experiments without any fine-tuning; we also analyse the impact of the network depth in Sect.~\ref{s:results}.

\paragraph{Optimisation}
For the \textsc{ar} specialisation step, we adopt the original suggested model setup. Hyperparameter values are set to: $\delta_{att}=0.6$, $\delta_{rep}=0.0$, $\lambda_{reg}=10^{-9}$ \cite{Mrksic:2017tacl}. The models are trained for 5 epochs with Adagrad \cite{Duchi:11}, with batch sizes set to $b_{att}=b_{rep}=50$, again as in the original work. 

For training the non-linear mapping with DFFN (Fig.~\ref{fig:main2}), we use the Adam algorithm \cite{Kingma:2015iclr} with default settings. The model is trained for 100 epochs with early stopping on a validation set. We reserve 10\% of all available seen data (i.e., the words from $\mathcal{V}_s$ represented in $\mathbf{X}_s$ and $\mathbf{X}'_s$) for validation, the rest are used for training. For the \textsc{mm} objective, we set $\delta_{mm}=0.6$ and $k=25$ in all experiments without any fine-tuning.

\section{Results and Discussion}
\label{s:results}
\begin{table*}[t]
\centering
\def\arraystretch{0.93}
\vspace{-0.0em}
{\footnotesize
\begin{tabularx}{\linewidth}{l ll ll ll ll ll ll}
\toprule
{} & \multicolumn{6}{c}{Setup: \textit{hold-out}} & \multicolumn{6}{c}{Setup: \textit{all}} \\
\cmidrule(lr){2-7} \cmidrule(lr){8-13}
{} & \multicolumn{2}{l}{\textsc{glove}} & \multicolumn{2}{l}{\textsc{sgns-bow2}} & \multicolumn{2}{l}{\textsc{fastText}} & \multicolumn{2}{l}{\textsc{glove}} & \multicolumn{2}{l}{\textsc{sgns-bow2}} & \multicolumn{2}{l}{\textsc{fastText}} \\
\cmidrule(lr){2-7} \cmidrule(lr){8-13}
{} & {SL} & {SV} & {SL} & {SV} & {SL} & {SV} & {SL} & {SV} & {SL} & {SV} & {SL} & {SV} \\
\cmidrule(lr){2-3} \cmidrule(lr){4-5} \cmidrule(lr){6-7} \cmidrule(lr){8-9} \cmidrule(lr){10-11} \cmidrule(lr){12-13}
{\textbf{Distributional:} $\mathbf{X}_d$} & {.408} & {.286} & {.414} & {.275} & {.383} & {.255} & {.408} & {.286} & {.414} & {.275} & {.383} & {.255} \\
{\textbf{+\textsc{ar} specialisation:} $\mathbf{X}'_s$} & {.408} & {.286} & {.414} & {.275} & {.383} & {.255} & {.690} & {.578} & {.658} & {.544} & {.629} & {.502} \\
{\textbf{++Mapping unseen:} $\mathbf{X}_f$} & {} & {} & {} & {} & {} & {} & {} & {} & {} & {} & {} & {} \\
{\textsc{linear-mse}} & {.504} & {.384} & {.447} & {.309} & {.405} & {.285} & {.690} & {.578} & {.656} & {.551} & {.628} & {.502} \\
{\textsc{nonlinear-mse}} & {.549} & {.407} & {.484} & {.344} & {.459} & {.329} & {.694} & {.586} & {.663} & {.556} & {.631} & {.506} \\
{\textsc{linear-mm}} & {.548} & {.422} & {.468} & {.329} & {.419} & {.308} & {.697} & {.582} & {.663} & {.554} & {.628} & {.487} \\
{\textsc{nonlinear-mm}} & {\bf .603} & {\bf .480} & {\bf .531} & {\bf .391} & {\bf .471} & {\bf .349} & {\bf .705} & {\bf .600} & {\bf .667} & {\bf .562} & {\bf .638} & {\bf .507} \\

\bottomrule
\end{tabularx}
}
%\vspace{-0.5em}
\caption{Spearman's $\rho$ correlation scores for three word vector collections on two English word similarity datasets, SimLex-999 (SL) and SimVerb-3500 (SV), using different mapping variants, evaluation protocols, and word vector spaces: from the initial distributional space $\mathbf{X}_d$ to the fully specialised space $\mathbf{X}_f$. $H=5$.}
%\vspace{-0.5mm}
\label{tab:11vectors}
\end{table*}

\begin{figure*}[t]
    \centering
    \begin{subfigure}[t]{0.32\linewidth}
        \centering
        \includegraphics[width=0.99\linewidth]{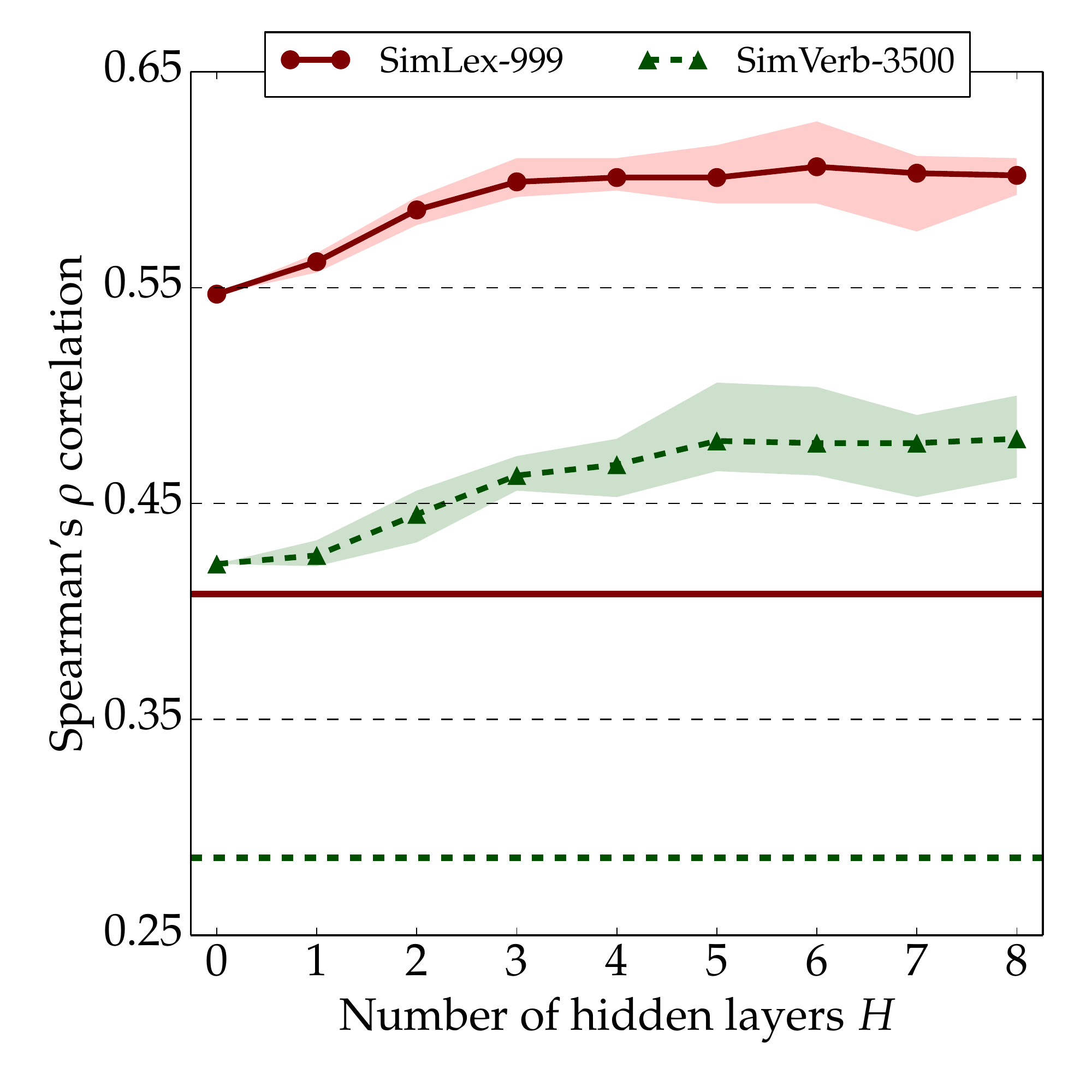}
        %\vspace{-0.7em}
        \caption{\textsc{glove}}
        \label{fig:glove}
    \end{subfigure}
    \begin{subfigure}[t]{0.32\textwidth}
        \centering
        \includegraphics[width=0.99\linewidth]{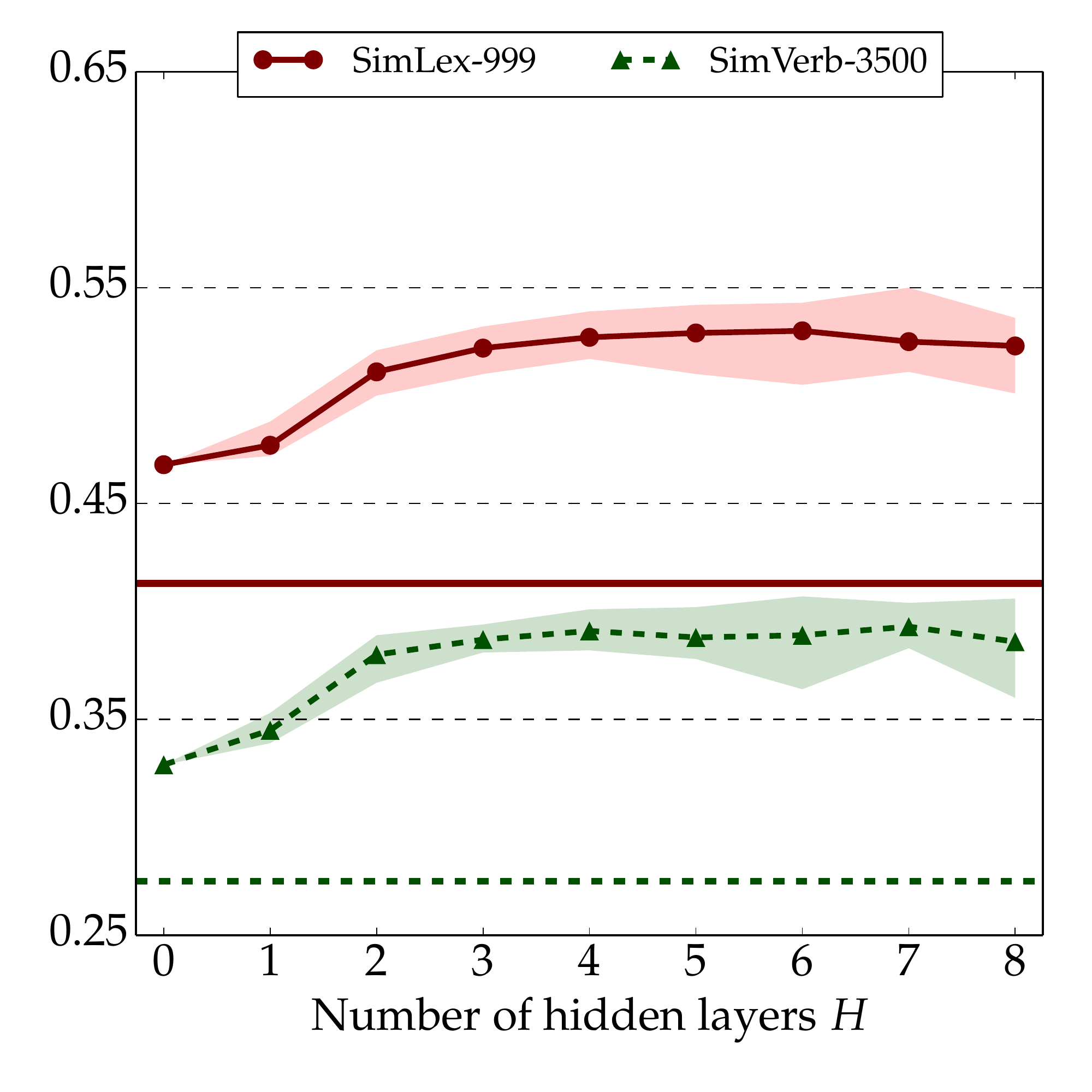}
        %\vspace{-0.7em}
        \caption{\textsc{sgns-bow2}}
        \label{fig:bow2}
    \end{subfigure}
    \begin{subfigure}[t]{0.32\textwidth}
        \centering
        \includegraphics[width=0.99\linewidth]{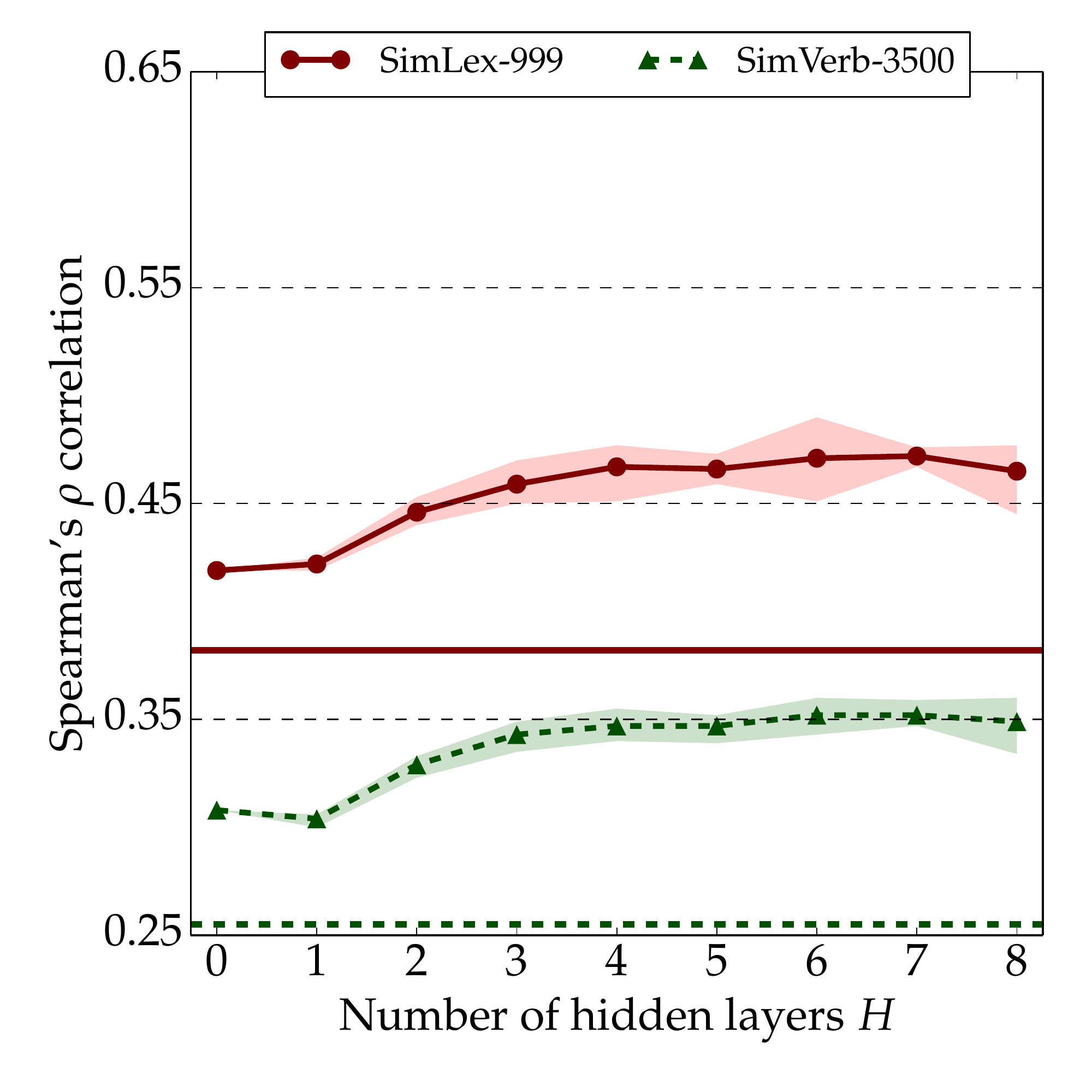}
        %\vspace{-0.7em}
        \caption{\textsc{fastText}}
        \label{fig:ft}
    \end{subfigure}
    %\vspace{-1.5mm}
    \caption{The results of the \textit{hold-out} experiments on SimLex-999 and SimVerb-3500 after applying our non-linear vector space transformation with different depths (hidden layer size $H$, see Fig.~\ref{fig:main2}). The results are presented as averages over 20 runs with the \textsc{nonlinear-mm} variant, the shaded regions are spanned by the maximum and minimum scores obtained. Thick horizontal lines refer to Spearman's rank correlations achieved in the initial space $\mathbf{X}_d$. $H=0$ denotes the standard \textit{linear} regression model \cite{Mikolov:2013arxiv,Lazaridou:2015acl} (\textsc{linear-mm} shown since it outperforms \textsc{linear-mse}).}
%\vspace{-0.5mm}
\label{fig:perlayer}
\end{figure*}

%Removed from the caption: All words present in SimLex and SimVerb were removed from the input linguistic constraints for the initial \textsc{attract-repel} specialisation model (i.e., they are in the vocabulary $\mathcal{V}_u$): this leaves the scores unaffected after the specialisation process (see Fig.~\ref{fig:main1})

\subsection{Intrinsic Evaluation: Word Similarity}
\label{ss:intrinsic}
\paragraph{Evaluation Protocol} 
The first set of experiments evaluates vector spaces with different specialisation procedures intrinsically on word similarity benchmarks: we use the SimLex-999 dataset \cite{Hill:2015cl}, and SimVerb-3500 \cite{Gerz:2016emnlp}, a recent verb pair similarity dataset providing similarity ratings for 3,500 verb pairs.\footnote{While other gold standards such as WordSim-353 \cite{Finkelstein:2002tois} or MEN \cite{Bruni:2014jair} coalesce the notions of true semantic similarity and (more broad) conceptual relatedness, SimLex and SimVerb provide explicit guidelines to discern between the two, so that related but non-similar words (e.g. \textit{tiger} and \textit{jungle}) have a low rating.} Spearman’s $\rho$ rank correlation is used as the evaluation metric.

We evaluate word vectors in two settings. First, in a synthetic \textit{hold-out} setting, we remove all linguistic constraints which contain words from the SimLex \textit{and} SimVerb evaluation data, effectively forcing all SimLex and SimVerb words to be \textit{unseen} by the \textsc{ar} specialisation model. The specialised vectors for these words are estimated by the learned non-linear DFFN mapping model. Second, the \textit{all} setting is a standard ``real-life'' scenario where some test (SimLex/SimVerb) words do occur in the constraints, while the mapping is learned for the remaining words.

\paragraph{Results and Analysis}
The results with the three word vector collections are provided in Tab.~\ref{tab:11vectors}. In addition, Fig.~\ref{fig:perlayer} plots the influence of the network depth $H$ on the model's performance. %We can draw several interesting conclusions. 

The results suggest that the mapping of unseen words is universally useful, as the highest correlation scores are obtained with the final fully specialised vector space $\mathbf{X}_f$ for all three input spaces. The results in the \textit{hold-out} setup are particularly indicative of the improvement achieved by our post-specialisation method. For instance, it achieves a +0.2 correlation gain with \textsc{glove} on both SimLex and SimVerb by specialising vector representations for words present in these datasets \textit{without} seeing a single external constraint which contains any of these words. This suggests that the perturbation of the seen subspace $\mathbf{X}_s$ by \textsc{attract-repel} contains implicit knowledge that can be propagated to $\mathbf{X}_u$, learning better representations for unseen words. We observe small but consistent improvements across the board in the \textit{all} setup. The smaller gains can be explained by the fact that a majority of SimLex and SimVerb words are present in the external constraints (93.7\% and 87.2\%, respectively).

The scores also indicate that \textit{both} non-linearity and the chosen objective function contribute to the quality of the learned mapping: largest gains are reported with the \textsc{nonlinear-mm} variant which \textbf{a)} employs non-linear activations and \textbf{b)} replaces the basic mean-squared-error objective with max-margin. The usefulness of the latter has been established in prior work on cross-space mapping learning \cite{Lazaridou:2015acl}. The former indicates that the initial \textsc{ar} transformation is non-linear. It is guided by a large number of constraints; their effect cannot be captured by a simple linear map as in prior work on, e.g., cross-lingual word embeddings \cite{Mikolov:2013arxiv,Ruder:2017arxiv}.

Finally, the analysis of the network depth $H$ indicates that going deeper helps only to a certain extent. Adding more layers allows for a richer parametrisation of the network (which is beneficial given the number of linguistic constraints used by \textsc{ar}). This makes the model more expressive, but it seems to saturate with larger $H$ values.
\begin{figure}[b]
    \centering
	\includegraphics[width=0.94\linewidth]{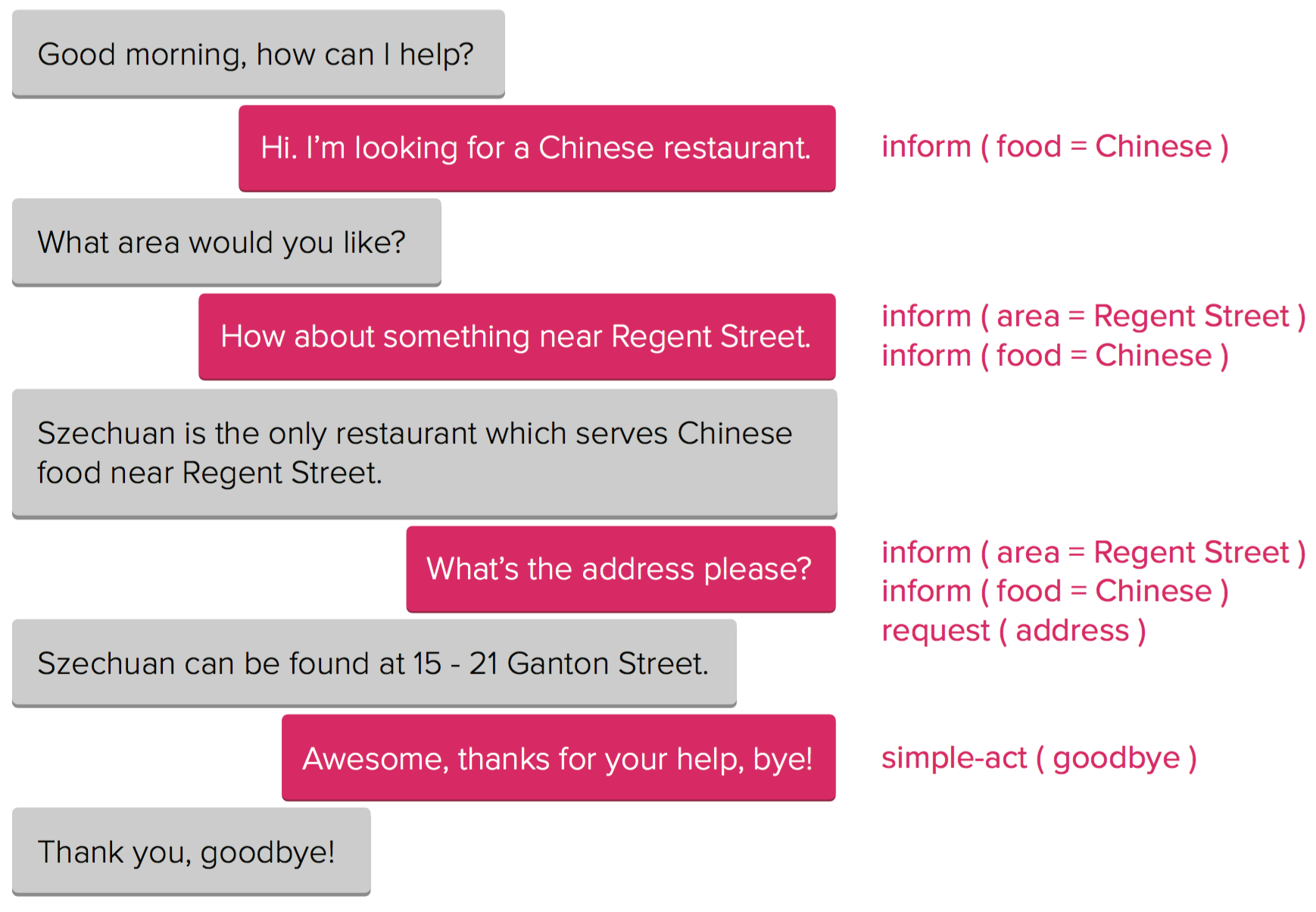}
    \vspace{-1mm}
    \caption{DST labels (user goals given by slot-value pairs) in a multi-turn dialogue \cite{Mrksic:15}. }%User goals are given by sets of constraints expressed by slot-value pairs.}
\vspace{0mm}
\label{fig:dst}
\end{figure}

\paragraph{Post-Specialisation with Other Post-Processors} We also verify that our post-specialisation approach is not tied to the \textsc{attract-repel} method, and is indeed applicable on top of any post-processing specialisation method. We analyse the impact of post-specialisation in the \textit{hold-out} setting using the original \textit{retrofitting} (\textit{RFit}) model \cite{Faruqui:2015naacl} and \textit{counter-fitting} (\textit{CFit}) \cite{Mrksic:2016naacl} in lieu of \text{attract-repel}. The results on word similarity with the best-performing \textsc{nonlinear-mm} variant are summarised in Tab.~\ref{tab:otherp}. 

The scores again indicate the usefulness of post-specialisation. As expected, the gains are lower than with \textsc{attract-repel}. \textit{RFit} falls short of \textit{CFit} as by design it can leverage only synonymy (i.e., \textsc{attract}) external constraints.

\begin{table}[t]
\centering
\def\arraystretch{0.93}
\vspace{-0.0em}
{\footnotesize
\begin{tabularx}{\linewidth}{l ll ll ll}
\toprule
%{} & \multicolumn{6}{c}{Setup: \textit{held-out}} \\
%\cmidrule(lr){2-7} 
{} & \multicolumn{2}{l}{\textsc{glove}} & \multicolumn{2}{l}{\textsc{sgns-bow2}} & \multicolumn{2}{l}{\textsc{fastText}}  \\
\cmidrule(lr){2-7} 
{} & {SL} & {SV} & {SL} & {SV} & {SL} & {SV} \\
\cmidrule(lr){2-3} \cmidrule(lr){4-5} \cmidrule(lr){6-7}
{$\mathbf{X}_d$} & {.408} & {.286}  & {.414} & {.275}  & {.383} & {.255} \\
{$\mathbf{X}_f$: \textit{RFit}} & {.493} & {.365}  & {.412} & {.285}  & {.413} & {.279} \\
{$\mathbf{X}_f$: \textit{CFit}} & {.540} & {.401}  & {.439} & {.318}  & {.306} & {.441} \\

\bottomrule
\end{tabularx}
}
\vspace{-0.5em}
\caption{Post-specialisation applied to two other post-processing methods. SL: SimLex; SV: SimVerb. \textit{Hold-out} setting. \textsc{nonlinear-mm}.}
%\vspace{-0.5mm}
\label{tab:otherp}
\end{table}

%\paragraph{Post-Specialisation vs. Direct Mapping} 
%Post-specialisation learns the non-linear specialisation function for unseen words in a post-hoc manner from the word embeddings fine-tuned using external constraints. One could also learn the mapping by \textit{directly} using external constraints as training instances. However, our results reveal that the direct mapping approach falls short of post-specialisation, even when relying on a deep network architecture: we see a decrease in performance for all three input vector spaces (e.g., $\approx$ 5 $\rho$ points on average on both SimLex and SimVerb in the \textit{hold-out} setup).\footnote{As the focus of the paper is on post-specialisation, we provide a detailed description of the direct mapping model along with the quantitative comparison as supplementary material.}

\subsection{Downstream Task I: DST}
\label{ss:dst}

Next, we evaluate the usefulness of post-specialisation for two downstream tasks -- dialogue state tracking and lexical text simplification -- in which discerning semantic similarity from other types of semantic relatedness is crucial. We first evaluate the importance of post-specialisation for a downstream language understanding task of \textit{dialogue state tracking} (DST) \cite{Henderson:14a,Williams:16}, adopting the evaluation protocol and data of \newcite{Mrksic:2017tacl}.

%\subsubsection{Dialogue State Tracking}

\paragraph{DST: Model and Evaluation}
The DST model is the first component of modern dialogue pipelines \cite{young:10}, which captures the users' goals at each dialogue turn and then updates the dialogue state. Goals are represented as sets of constraints expressed as slot-value pairs (e.g., food=\textit{Chinese}). The set of
slots and the set of values for each slot constitute
the ontology of a dialogue domain. The probability distribution over the possible states is the system's estimate of the user's goals, and it is used by the dialogue manager module to select the subsequent system response \cite{su:2016:nnpolicy}. An example in Fig.~\ref{fig:dst} illustrates the DST pipeline.

For evaluation, we use the Neural Belief Tracker (NBT), a state-of-the-art DST model which was the first to reason purely over pre-trained word vectors \cite{Mrksic:2017acl}.\footnote{https://github.com/nmrksic/neural-belief-tracker} The NBT uses no hand-crafted semantic lexicons, instead composing word vectors into intermediate utterance and context representations.\footnote{The NBT keeps word vectors \textit{fixed} during training to enable generalisation for words unseen in DST training data.} For full model details, we refer the reader to the original paper. The importance of word vector specialisation for the DST task (e.g., distinguishing between synonyms and antonyms by pulling \textit{northern} and \textit{north} closer in the vector space while pushing \textit{north} and \textit{south} away) has been established \cite{Mrksic:2017acl}.

\begin{table}[t]
\centering
\def\arraystretch{0.95}
\vspace{-0.0em}
{\footnotesize
\begin{tabularx}{\linewidth}{l XX}
\toprule
{\textsc{English}} & \textit{hold-out} & \textit{all} \\
\cmidrule(lr){2-2} \cmidrule(lr){3-3}
%\textbf{Random-Init} of $\mathcal{V}_d$ & {.809} & {.809} \\
%\midrule
\textbf{Distributional: $\mathbf{X}_d$} & {.797} & {.797} \\
\textbf{+\textsc{ar} Spec.: $\mathbf{X}'_s \cup \mathbf{X}_u$} & {.797} & {.817} \\
\textbf{++Mapping: $\mathbf{X}_f = \mathbf{X}'_s \cup \widehat{\mathbf{X}'_u}$} & {} & {} \\
\textsc{linear-mm} & {.815} & {.818} \\
\textsc{nonlinear-mm} & {\bf .827} & {\bf.835} \\

\bottomrule
\end{tabularx}
}
\vspace{-0.5em}
\caption{DST results in two evaluation settings (\textit{hold-out} and \textit{all}) with different \textsc{glove} variants.}
%\vspace{-1.5mm}
\label{tab:endst}
\end{table}

Again, as in prior work the DST evaluation is based on the Wizard-of-Oz (WOZ) v2.0 dataset \cite{Wen:17,Mrksic:2017acl}, comprising 1,200 dialogues split into training (600 dialogues), development (200), and test data (400). In all experiments, we report the standard DST performance measure: \textit{joint goal accuracy}, and report scores as averages over 5 NBT training runs.

% (IV Removed): The entire dataset has also been translated to German and Italian \cite{Mrksic:2017tacl}. 
\begin{table*}[t]
\centering
\vspace{-0.0em}
{\footnotesize
\begin{tabularx}{\linewidth}{l XXXX XXXX}
\toprule
{} & \multicolumn{4}{X}{\textsc{German}} & \multicolumn{4}{X}{\textsc{Italian}} \\
\cmidrule(lr){2-5} \cmidrule(lr){6-9}
{} & \multicolumn{2}{l}{SimLex (Similarity)} & \multicolumn{2}{l}{WOZ (DST)} & \multicolumn{2}{l}{SimLex (Similarity)} & \multicolumn{2}{l}{WOZ (DST)} \\
{} & {\textit{hold-out}} & {\textit{all}} & {\textit{hold-out}} & {\textit{all}} & {\textit{hold-out}} & {\textit{all}} & {\textit{hold-out}} & {\textit{all}}  \\
\cmidrule(lr){2-3} \cmidrule(lr){4-5} \cmidrule(lr){6-7} \cmidrule(lr){8-9}
\textbf{Distributional: $\mathbf{X}_d$} & {.267} & {.267} & {.487} & {.487} & {.363} & {.363} & {.618} & {.618} \\
\textbf{+\textsc{ar} Spec.: $\mathbf{X}'_s \cup \mathbf{X}_u$} & {.267} & {.422} & {.487} & {.535} & {.363} & {\bf .616} & {.618} & {.634} \\
\textbf{++Mapping: $\mathbf{X}_f$} & {} & {} & {} & {} & {} & {} & {} & {} \\
\textsc{linear-mm} & {.354} & {.449} & {.485} & {.533} & {.401} & {\bf .616} & {.627} & {.633} \\
\textsc{nonlinear-mm} & {\bf .367} & {\bf .466} & {\bf .496} & {\bf .538} & {\bf .428} & {\bf .616} & {\bf .637} & {\bf .647} \\
\bottomrule
\end{tabularx}
}
%\vspace{-0.5em}
\caption{Results on word similarity (Spearman's $\rho$) and DST (joint goal accuracy) for German and Italian.}
%\vspace{-1.5mm}
\label{tab:deitdst}
\end{table*}
\paragraph{Results and Analysis}
We again evaluate word vectors in two settings: \textbf{1)} \textit{hold-out}, where linguistic constraints with words appearing in the WOZ data are removed, making all WOZ words unseen by \textsc{attract-repel}; and \textbf{2)} \textit{all}. The results for the English DST task with different \textsc{glove} word vector variants are summarised in Tab.~\ref{tab:endst}; similar trends in results are observed with two other word vector collections. The scores maintain conclusions established in the word similarity task. First, semantic specialisation with \textsc{attract-repel} is again beneficial, and discerning between synonyms and antonyms improves DST performance. However, specialising unseen words (the final $\mathbf{X}_u$ vector space) yields further improvements in both evaluation settings, supporting our claim that the specialisation signal can be propagated to unseen words. % unseen in the constraints.

This downstream evaluation again demonstrates the importance of non-linearity, as the peak scores are reported with the \textsc{nonlinear-mm} variant. More substantial gains in the \textit{all} setup are observed in the DST task compared to the word similarity task. This stems from a lower coverage of the WOZ data in the \textsc{ar} constraints: 36.3\% of all WOZ words are unseen words. Finally, the scores are higher on average in the \textit{all} setup, since this setup uses more external constraints for \textsc{ar}, and consequently uses more training examples to learn the mapping.

\paragraph{Other Languages} 
We test the portability of our framework to two other languages for which we have similar evaluation data: German (DE) and Italian (IT). SimLex-999 has been translated and rescored in the two languages by \newcite{Leviant:2015arxiv}, and the WOZ data were translated and adapted by \newcite{Mrksic:2017tacl}. Exactly the same setup is used as in our English experiments, without any additional language-specific fine-tuning. Linguistic constraints were extracted from the same sources: synonyms from the PPDB (135,868 in DE, 362,452 in IT), antonyms from BabelNet (4,124 in DE, and 16,854 in IT). Our starting distributional vector spaces are taken from prior work: IT vectors are from \cite{Dinu:2015arxiv}, DE vectors are from \cite{Vulic:2016acluniversal}. The results are summarised in Tab.~\ref{tab:deitdst}.

Our post-specialisation approach yields consistent improvements over the initial distributional space and the \textsc{ar} specialisation model in both tasks and for both languages. We do not observe any gain on IT SimLex in the \textit{all} setup since IT constraints have almost complete coverage of all IT SimLex words (99.3\%; the coverage is 64.8\% in German). As expected, the DST scores in the \textit{all} setup are higher than in the \textit{hold-out} setup due to a larger number of constraints and training examples.

Lower absolute scores for Italian and German compared to the ones reported for English are due to multiple factors, as discussed previously by \newcite{Mrksic:2017tacl}: \textbf{1)} the \textsc{ar} model uses less linguistic constraints for DE and IT; \textbf{2)} distributional word vectors are induced from smaller corpora; \textbf{3)}  linguistic phenomena (e.g., cases and compounding in DE) contribute to data sparsity and also make the DST task more challenging. However, it is important to stress the consistent gains over the vector space specialised by the state-of-the-art \textsc{attract-repel} model across all three test languages. This indicates that the proposed approach is language-agnostic and portable to multiple languages. 

\subsection{Downstream Task II: Lexical Simplification}
\label{ss:lts}

In our second downstream task, we examine the effects of post-specialisation on lexical simplification (LS) in English. LS aims to substitute complex words (i.e., less commonly used words) with their simpler synonyms in the context. Simplified text must keep the meaning of the original text, which is discerning similarity from relatedness is important (e.g., in \textit{``The automobile was set on fire''} the word \textit{``automobile''} should be replaced with \textit{``car''} or \textit{``vehicle''} but not with \textit{``wheel''} or \textit{``driver''}). 
\setlength{\tabcolsep}{7pt}
\begin{table}[t]
\centering
\vspace{-0.0em}
{\footnotesize
\begin{tabularx}{\linewidth}{l l c c}
\toprule
Vectors & Specialisation & Acc. & Ch. \\ \midrule
 & \textbf{Distributional: $\mathbf{X}_d$} & {66.0} & \textbf{94.0} \\
\textsc{Glove} & \textbf{+\textsc{ar} Spec.: $\mathbf{X}'_s \cup \mathbf{X}_u$} & {67.6} & {87.0} \\ 
& \textbf{++Mapping: $\mathbf{X}_f$} & \textbf{72.3} & {87.6} \\ \midrule
& \textbf{Distributional: $\mathbf{X}_d$} & {57.8} & {84.0} \\
\textsc{fastText} & \textbf{+\textsc{ar} Spec.: $\mathbf{X}'_s \cup \mathbf{X}_u$} & {69.8} & \textbf{89.4} \\ 
& \textbf{++Mapping: $\mathbf{X}_f$} & \textbf{74.3} & {88.8} \\ \midrule
& \textbf{Distributional: $\mathbf{X}_d$} & {56.0} & {79.1} \\
\textsc{sgns-bow2} & \textbf{+\textsc{ar} Spec.: $\mathbf{X}'_s \cup \mathbf{X}_u$} & {64.4} & {86.7} \\ 
& \textbf{++Mapping: $\mathbf{X}_f$} & \textbf{70.9} & \textbf{86.8} \\
\bottomrule
\end{tabularx}}
\vspace{-0.5mm}
\caption{Lexical simplification performance with post-specialisation applied on three input spaces.}
\label{tbl:simplification}
\end{table}
%

%\paragraph{Model and Evaluation} 
We employ \textsc{Light-LS} \cite{glavavs-vstajner:2015:ACL-IJCNLP}, a lexical simplification algorithm that: 1) makes substitutions based on word similarities in a semantic vector space, and 2) can be provided an arbitrary embedding space as input.\footnote{\url{https://github.com/codogogo/lightls}} For a complex word, \textsc{Light-LS} considers the most similar words from the vector space as simplification candidates. Candidates are ranked according to several features, indicating simplicity and fitness for the context (semantic relatedness to the context of the complex word). The substitution is made if the best candidate is simpler than the original word. By providing vector spaces post-specialised for semantic similarity to \textsc{Light-LS}, we expect to more often replace complex words with their true synonyms.

We evaluate \textsc{Light-LS }performance in the \textit{all} setup on the LS benchmark compiled by \newcite{horn2014learning}, who crowdsourced 50 manual simplifications for each complex word.
As in prior work, we evaluate performance with the following metrics: 1) 
%\textit{Precision} (P) is the percentage of correct simplifications, i.e., the percentage of cases in which the system's replacement is found among the crowdsourced substitutions. Precision does not consider (i.e., does not penalize the system for) cases in which the system did not simplify the complex word at all; 
\textit{Accurracy} (Acc.) is the number of correct simplifications made (i.e., the system made the simplification and its substitution is found in the list of crowdsourced substitutions), divided by the total number of indicated complex words; 2) \textit{Changed} (Ch.) is the percentage of indicated complex words that were replaced by the system (whether or not the replacement was correct).
%Unlike \textit{Precision}, \textit{Accuracy} penalizes the system also for not making a simplification when it should have -- i.e., in a sense, \textit{Accuracy} is a combination of \textit{Precision} and \textit{Changed}. 
%

%\paragraph{Results} 

LS results are summarised in Tab.~\ref{tbl:simplification}. Post-specialised vector spaces consistently yield 5-6\% gain in \textit{Accuracy} compared to respective distributional vectors and embeddings specialised with the state-of-the-art \textsc{Attract-Repel} model. Similar to DST evaluation, improvements over \textsc{Attract-Repel} demonstrate the importance of specialising the vectors of the entire vocabulary and not only the vectors of words from the external constraints.

\section{Conclusion and Future Work}
\label{s:conclusion}

We have presented a novel post-processing model, termed \emph{post-specialisation}, that specialises \iffalse (i.e., updates) \fi word vectors for the full vocabulary of the input vector space. Previous post-processing specialisation models fine-tune word vectors only for words occurring in external lexical resources. \iffalse (e.g., WordNet, BabelNet, PPDB). \fi In this work, we have demonstrated that the specialisation of the subspace of seen words can be leveraged to learn a mapping function which specialises vectors for all other words, unseen in the external resources. Our results across word similarity and downstream language understanding tasks show consistent improvements over the state-of-the-art specialisation method for all three test languages.

In future work, we plan to extend our approach to specialisation for asymmetric relations such as hypernymy or meronymy \cite{Glavas:2017emnlp,Nickel:2017arxiv,Vulic:2018lear}. We will also investigate more sophisticated non-linear functions. \iffalse For instance, instead of a single global mapping function, already proven effective in this work, one may also learn multiple local mapping functions for different vector space regions based on a pre-mapping clustering step. \fi The code is available at: \url{https://github.com/cambridgeltl/post-specialisation/}.

\section*{Acknowledgments}
We thank the three anonymous reviewers for their insightful suggestions. This work is supported by the ERC Consolidator Grant LEXICAL: Lexical Acquisition Across Languages (no 648909).

\bibliography{acl2017_refs}
\bibliographystyle{acl_natbib}

%\clearpage
%\input{supplement}

\end{document}